\documentclass[10pt, a4paper]{article}

\usepackage[final]{lrec2026} 
\usepackage{multirow} 
\usepackage{booktabs}
\usepackage{amsmath}
\usepackage{array}
\usepackage{makecell}
\usepackage{tcolorbox}
\usepackage{xcolor} 
\usepackage[T1]{fontenc}
\usepackage{enumitem}

\title{CGU-ILALab at FoodBench-QA 2026: Comparing Traditional and LLM-based Approaches for Recipe Nutrient Estimation}

\name{Wei-Chun Chen$^a$, Yu-Xuan Chen$^b$, I-Fang Chung$^a$, Ying-Jia Lin$^{c,d}$}

\address{$^a$Institute of Biomedical Informatics, National Yang Ming Chiao Tung University, Taipei, Taiwan \\
         $^b$Institute of Health Data Science, Chang Gung University, Taoyuan, Taiwan \\
         $^c$Department of Artificial Intelligence, Chang Gung University, Taoyuan, Taiwan \\
         $^d$Artificial Intelligence Research Center, Chang Gung University, Taoyuan, Taiwan \\
         \{andybox11111.md14, ifchung\}@nycu.edu.tw, \{m1462004, yjlin\}@cgu.edu.tw}

\abstract{
Accurate nutrient estimation from unstructured recipe text is an important yet challenging problem in dietary monitoring, due to ambiguous ingredient terminology and highly variable quantity expressions.
We systematically evaluate models spanning a wide range of representational capacity, from lexical matching methods (TF-IDF with Ridge Regression), to deep semantic encoders (DeBERTa-v3), to generative reasoning with large language models (LLMs).
Under the strict tolerance criteria defined by EU Regulation 1169/2011, our empirical results reveal a clear trade-off between predictive accuracy and computational efficiency.
The TF-IDF baseline achieves moderate nutrient estimation performance with near-instantaneous inference, whereas the DeBERTa-v3 encoder performs poorly under task-specific data scarcity.
In contrast, few-shot LLM inference (e.g., Gemini 2.5 Flash) and a hybrid LLM refinement pipeline (TF-IDF combined with Gemini 2.5 Flash) deliver the highest validation accuracy across all nutrient categories.
These improvements likely arise from the ability of LLMs to leverage pre-trained world knowledge to resolve ambiguous terminology and normalize non-standard units, which remain difficult for purely lexical approaches.
However, these gains come at the cost of substantially higher inference latency, highlighting a practical deployment trade-off between real-time efficiency and nutritional precision in dietary monitoring systems.
\\ \newline \Keywords{Recipe Nutrient Estimation, Large Language Models} }

\begin{document}

\maketitleabstract

\section{Introduction}

Accurate nutrient estimation from recipe text is a practically relevant yet technically challenging task in the food and nutrition domain. While large language models (LLMs) have demonstrated remarkable capabilities in general-purpose reasoning and knowledge retrieval, it remains unclear how well they handle the structured, quantitative reasoning required for food applications. Recipe texts present unique difficulties: ambiguous ingredient terminology, non-standard measurement units, and the need to map diverse culinary expressions to precise nutritional values. These challenges make recipe nutrient estimation a compelling testbed for evaluating whether different modeling paradigms --- from traditional lexical methods to LLM-based inference --- can produce reliable nutritional predictions.

The FoodBench-QA 2026 shared task\footnote{\url{https://www.codabench.org/competitions/12112/}} asks systems to answer food and nutrition questions based on unstructured recipe text.
The complexity of the food domain stems from its inherent heterogeneity. Unlike standardized datasets in other fields, culinary data is characterized by ambiguous ingredient terminology, non-standard and highly variable measurement units (e.g., ``a pinch,'' ``a medium bunch''), and the challenge of mapping these varied expressions to structured food composition databases. A model must not only perform accurate entity disambiguation to distinguish between semantically similar but nutritionally distinct ingredients (e.g., ``coconut milk'' vs. ``coconut water'') but also execute complex reasoning to normalize quantities and aggregate nutritional values from disparate sources.

\begin{table}[ht]
    \centering
    \begin{tabular}{p{7.27cm}}
        \toprule
        \textbf{Case 1 Prompt:} Assess the nutritional profile of 180 ml of pineapple juice, canned or bottled, unsweetened, without added ascorbic acid. \\
        \textbf{Case 1 Answer:} The nutrient values demonstrated here are: energy - 99.79, fat - 0.23, protein - 0.68, salt - 3.78, saturates - 0.02, sugars - 18.86. \\
        \midrule
        \textbf{Case 2 Prompt:} Check the nutritional values per 100 g in a recipe that comprises these ingredients: 2 teaspoons corn, sweet, white, raw, 2 teaspoons corn, sweet, white, raw, 2 teaspoons peppers, sweet, green, raw, 1 teaspoon spices, coriander seed, 1/2 cup butter, without salt. \\
        \textbf{Case 2 Answer:} Nutrient details in 100 g: energy - 713.49, fat - 80.53, protein - 0.86, salt - 0.04, saturates - 50.10, sugars - 0.09. \\
        \bottomrule
    \end{tabular}
    \caption{Examples in the FoodBench-QA 2026 shared task. Prompt: nutritional queries; Answer: response that should be provided by a system.}
    \label{tab:examples}
\end{table}

In response to these challenges, the FoodBench-QA 2026 shared task calls for systems capable of moving beyond simple pattern matching to provide accurate nutritional estimates that comply with regulatory standards.
Motivated by this critical need, this paper focuses on the sub-task of nutrient estimation for recipes, 
where examples from this sub-task are shown in Table \ref{tab:examples}.
Though each data label includes six target nutrients (Energy, Salt, Fat, Protein, Saturates, Sugars), the FoodBench-QA shared task focused on evaluation of the latter four nutrients.
To tackle the ``real-world'' noise of dietary data, we systematically explore and evaluate multiple methodologies.
By investigating various approaches to resolving ambiguity and standardizing measurements, we aim to identify the most effective strategies\footnote{Our code: \url{https://github.com/ila-lab/FoodBench-QA}} for improving nutrient estimation accuracy, thereby advancing the reliability of automated systems in health-oriented language technologies.

\section{Dataset}
The dataset utilized in this study is sourced from the ``FoodBench-QA: Shared Task on Grounded Food \& Nutrition Question Answering'' competition on Codabench.
Specifically, we employed the annotated subset of the data without titles (\textbf{T1.1 - Ingredients})\footnote{\url{https://github.com/matejMartinc/FoodBench-QA-train/tree/main/recipe\%20nutrient\%20assestment/ingredients}}.
During the preprocessing stage, we identified and removed duplicate recipe entries to ensure data quality.
After the data cleaning process, a total of 14,512 unique samples remained.
Finally, we partitioned the dataset into our training and validation sets at an 80:20 ratio, resulting in 11,609 and 2,903 examples for training and validation, respectively.

\section{Method}

To comprehensively investigate the applicability of models with varying complexities to this task, this study systematically compares multiple approaches ranging from traditional TF-IDF pipelines to LLM-based inference, as well as a hybrid strategy combining TF-IDF with an LLM.

Starting with fundamental lexical statistical methods, we implemented an approach combining TF-IDF \cite{salton1988tfidf} with traditional machine learning.
Subsequently, we advanced into the deep learning domain by employing an encoder-based model \cite{he2023debertav} to enhance the capability for semantic understanding.
Finally, we extended our research to current state-of-the-art technologies by directly utilizing large language models (LLMs) for few-shot inference. 
This design not only establishes a clear comparative baseline but also explicitly demonstrates the impact of technological evolution on overall task performance.

\subsection{Term Frequency-Inverse Document Frequency (TF-IDF) with Machine Learning} 

As our foundational baseline, we represent recipe texts as sparse lexical vectors using TF-IDF \cite{salton1988tfidf}, coupled with Ridge Regression \cite{hoerl1970ridge} for the continuous prediction tasks. TF-IDF is a widely adopted representation for converting text into numerical vectors in text-based prediction tasks \cite{jones1972statistical}.

TF-IDF features are constructed using recipe ingredient text as input, with basic preprocessing including lowercasing and the removal of English stop words at the word level. To capture both broader semantic information and subtle spelling variations inherent in culinary data, we extract two distinct types of TF-IDF features.

First, \textbf{Word-level TF-IDF} applies unigram and bigram features to capture key ingredient terms and common word combinations. To control feature quality and reduce noise, a minimum document frequency of $\text{min\_df} = 2$ is applied, retaining only terms that appear in at least two documents. A maximum document frequency of $\text{max\_df} = 0.9$ removes overly common terms appearing in more than 90\% of documents. The vocabulary size is limited to 8,000 features, and logarithmic scaling is applied to term frequencies to dampen the influence of high-frequency terms.

Second, following the character n-gram approach proposed by \citet{cavnar1994n}, \textbf{Char-level TF-IDF} utilizes 3--5 word-boundary-aware character n-grams to effectively handle spelling variations, typos, and fine-grained morphological patterns. For these features, $\text{min\_df} = 2$ and $\text{max\_df} = 0.95$ are applied, and the feature size is limited to 12,000 to balance expressiveness and dimensionality.

By combining these word-level and character-level features, we form a comprehensive final feature vector. This hybrid representation retains both word-level meaning and character-level details, making the model highly robust to variations in textual expression.

Furthermore, Ridge Regression, with its $L_2$ regularization, is particularly well-suited for handling the resulting high-dimensional and sparse feature space. Critically, it mitigates the multicollinearity naturally introduced by overlapping word and character features. By penalizing large coefficients, Ridge Regression prevents the model from over-concentrating its ``attention'' (i.e., assigning extreme weights) solely on perfectly matched words. Instead, it encourages a more balanced weight distribution, ensuring the model genuinely leverages character-level features to handle textual noise without overfitting.

Consequently, this methodology not only provides computational efficiency but also offers high interpretability, making it a strong baseline for ingredient terminology that can be matched explicitly.

\subsection{Encoder-based NLP Models} 

To overcome the limitations of sparse representations, we advance to deep context-aware architectures using pre-trained bidirectional encoders.
Specifically, we implement \textbf{DeBERTa-v3} (Decoding-enhanced BERT with disentangled attention) \cite{he2023debertav} as our core model for this stage. DeBERTa-v3 significantly improves upon standard BERT architectures \cite{devlin_bert_2019} by utilizing a disentangled attention mechanism, which encodes words and their relative positions separately. In principle, this architecture could benefit culinary data, where the relative positioning of numbers, units, and ingredients (e.g., ``2 tablespoons of olive oil'' vs. ``oil, 2 tablespoons'') carries important semantic information.

We directly feed the raw ingredient text into the model without additional preprocessing or manual extraction.
The model processes this sequence to generate context-aware embeddings before passing them to a linear regression head to predict nutritional values.

In theory, this deep learning tier could address the challenge of ambiguous ingredient terminology highlighted earlier.
By dynamically capturing the surrounding text and positional context, DeBERTa-v3 may facilitate implicit word sense disambiguation, potentially differentiating between semantically similar but nutritionally disparate items (e.g., inferring whether ``coconut'' refers to milk, water, or oil based on the recipe context). However, as we discuss in the results, these theoretical advantages did not materialize in our low-data setting.

\subsection{Large Language Models (LLMs): Direct Inference and Semantic Refinement} 

Finally, we extend our framework to the current state-of-the-art by utilizing generative LLMs. In this tier, we explore the capabilities of LLMs through two distinct operational paradigms: direct inference and post-processing refinement.

For \textbf{Direct Inference}, we evaluate a suite of locally deployed open-source LLMs, specifically GPT-OSS-20B \cite{openai2025gptoss}\footnote{\url{https://huggingface.co/openai/gpt-oss-20b}}, Gemma-3-27B \cite{gemmateam2025gemma3technicalreport}\footnote{\url{https://huggingface.co/google/gemma-3-27b-it}}, and Nemotron-3-Nano-30B \cite{nvidia2025nemotron3nanoopen}\footnote{\url{https://huggingface.co/nvidia/NVIDIA-Nemotron-3-Nano-30B-A3B-BF16}}, via few-shot prompting. In addition, we evaluate Gemini 2.5 Flash \cite{comanici2025gemini25} as a cloud-based direct inference model using the same few-shot prompting strategy, to compare API-served models against locally deployed ones.
The prompt design for direct inference, which includes two in-context examples (i.e., 2-shot), is shown in Figure \ref{fig:direct_prompt}.
Instead of traditional fine-tuning, we task these models with implicitly executing the entire estimation pipeline, encompassing ingredient extraction, unit normalization, and nutrient aggregation, within a single inference step. These models are particularly well-suited to address the persistent issue of non-standard and highly variable measurement units (e.g., rationally transforming ``a medium bunch'' or ``a pinch'' into quantifiable metrics). By leveraging their pre-trained world knowledge, they perform the multi-step reasoning required to bridge the gap between noisy culinary instructions and nutritional estimations.

\begin{figure}[h] 
    \centering
    \begin{tcolorbox}[
        colback=gray!5,      
        colframe=gray!60,    
        boxrule=0.8pt,       
        arc=2pt,             
        left=6pt, right=6pt, top=6pt, bottom=6pt 
    ]
    \small 
    \textbf{System Role:} \\
    You are a specialized nutritional data analyst. Your task is to calculate the nutrient profile per 100g for recipes provided in \texttt{[INST]} format.
    
    \vspace{0.5em} 
    
    \textbf{Instructions:}\\
        \textbf{Unit Conversion:} Convert all units (e.g., pounds, cups, tablespoons, ml) to grams (g) using standard conversion factors (e.g., 1 cup water $\approx$ 236.6g, 1 tablespoon butter $\approx$ 14.2g).\\
        \textbf{Calculation:} Sum the total weight and total nutrients of all ingredients, then normalize the values to a 100g portion.\\
        \textbf{Output Format:} You must only provide the final result in this specific format: \\
        \texttt{Nutrient values per 100 g: fat - [value], protein - [value], saturates - [value], sugars - [value]}
        \textbf{input1:} [INST] Identify the nutritional content per 100 grams for a recipe with the following ingredients: 2 tablespoon soy sauce made from soy (tamari), 1 tablespoon peanut butter, smooth style, without salt, ... [ingredients truncated for brevity, paper display only] [/INST]\\
        \textbf{answer1:} Nutrient values per 100 g: fat - 8.55, protein - 12.31, saturates - 1.72, sugars - 14.17
        \textbf{input2:} [INST] Identify the nutritional content per 100 grams for a recipe with the following ingredients: 1 cup wheat flour, 2 tbsp olive oil, ... [ingredients truncated for brevity, paper display only] [/INST]\\
        \textbf{answer2:} Nutrient values per 100 g: fat - 14.20, protein - 3.10, saturates - 2.15, sugars - 0.50
  
    \end{tcolorbox}
    \caption{The prompt design used for direct inference with an LLM.}
    \label{fig:direct_prompt}
\end{figure}

Complementing this direct approach, we introduce an \textbf{LLM-based Refinement} strategy to compensate for the limited semantic understanding of our shallower baseline models (e.g., the TF-IDF approach). For this post-processing step, we utilize Gemini 2.5 Flash \cite{comanici2025gemini25}, with the prompt shown in Figure \ref{fig:prompt_template}. For each test sample, the original ingredient text alongside the initial numerical predictions generated by the baseline model are provided as context to the LLM. The model evaluates whether the predicted values fall within reasonable nutritional boundaries and adjusts them when necessary. This step effectively acts as a semantic refinement layer, helping to align purely statistical or lexical predictions with real-world culinary logic and nutritional common sense.

\begin{figure}[h]
    \centering
    \begin{tcolorbox}[
        colback=gray!5,       
        colframe=blue!50,     
        boxrule=0.8pt,
        arc=2pt,
        left=6pt, right=6pt, top=6pt, bottom=6pt
    ]
    \small
    You are a nutrition expert.
    
    \vspace{0.5em}
    \textbf{Food:} \\
    \texttt{\{text\}} 
    
    \vspace{0.5em}
    \textbf{Predicted nutrients per 100g:} \\
    Protein: \texttt{\{pred['protein\_g']:.2f\}} \\ 
    Fat: \texttt{\{pred['fat\_g']:.2f\}} \\
    Sugar: \texttt{\{pred['sugars\_g']:.2f\}} \\
    Saturates: \texttt{\{pred['saturates\_g']:.2f\}}
    
    \vspace{0.5em}
    \textbf{Return JSON only with keys:} \\
    \texttt{protein\_g}, \texttt{fat\_g}, \texttt{sugars\_g}, \texttt{saturates\_g}
    \end{tcolorbox}
    \caption{The prompt template used to refine nutrient predictions. Variables in \texttt{monospaced} font are dynamically filled with an input query \texttt{(\{text\})} and TF-IDF prediction values during inference.}
    \label{fig:prompt_template}
\end{figure}

\section{Results}

In this section, we present the empirical results comparing multiple modeling approaches for recipe nutrient estimation.
We evaluate all approaches using the official evaluation tool, which operationalizes EU Regulation 1169/2011\cite{eu1169} tolerance thresholds as binary accuracy criteria — a more practically meaningful standard than continuous error metrics such as MAE or RMSE, as it directly reflects whether a model's predictions fall within the margins that regulators deem acceptable for food labeling.

\begin{table*}[t] 
\centering
\renewcommand\arraystretch{1.2} 
\begin{tabular}{lccccc}
\toprule
\multirow{2}{*}{\textbf{Model Architecture}} & \multicolumn{4}{c}{\textbf{Performance Metrics}} & \multirow{2}{*}{\textbf{Latency}} \\
\cmidrule{2-5} 
& \textbf{Sugar} & \textbf{Protein} & \textbf{Fat} & \textbf{Saturates} &  \\
\midrule
\multicolumn{6}{c}{\textit{Traditional Machine Learning}} \\
TF-IDF + Ridge Regression   & 50.26 & 64.62 & 40.58 & 50.33 & \textbf{1 ms} \\
\midrule
\multicolumn{6}{c}{\textit{Encoder-based model}} \\
DeBERTa-v3                  & 8.87  & 23.86  & 8.27  & 8.01  & 3.58 ms \\
\midrule
\multicolumn{6}{c}{\textit{LLM-based Direct Inference}} \\
GPT-OSS-20B     & 32.07 & 41.38 & 29.93 & 37.00 & 9.9 s \\
Gemma-3-27B        & 41.51 & 60.21 & 41.44 & 48.02 & 1.4 s \\
Nemotron-3-Nano-30B        & 22.05  & 30.58  & 19.73  & 26.23  & 23.7 s \\
Gemini 2.5 Flash  & 55.36 & 67.17 & \textbf{54.12} & \textbf{59.73} & 1.0 s \\
\midrule
\multicolumn{6}{c}{\textit{LLM-based Refinement}} \\
TF-IDF + Gemini 2.5 Flash & \textbf{55.46} & \textbf{67.79} & 52.12 & 58.39 & 1.0 s \\
\bottomrule
\end{tabular}
\caption{Performance in accuracy (\%) and computational cost of evaluated methods. Metrics follow EU Regulation 1169/2011 tolerance criteria. Bold indicates best performance per metric.}
\label{tab:comprehensive-results}
\end{table*}

\subsection{Performance Comparison}

As presented in Table \ref{tab:comprehensive-results}, our results reveal a consistent pattern: models with greater pre-trained world knowledge achieve higher accuracy, while lightweight approaches retain a decisive advantage in inference speed.

We first evaluated the TF-IDF and Ridge Regression pipeline as our baseline.
This approach achieved notable performance on Protein (64.62), while Fat (40.58), Saturates (50.33), and Sugar (50.26) remained at a moderate level. These results suggest that explicit lexical matching provides a reliable signal for nutrient estimation even without any semantic understanding.

To capture deeper semantic relationships, we subsequently evaluated the DeBERTa-v3 encoder. However, contrary to the typical dominance of deep encoder-based models in NLP tasks, DeBERTa-v3 performed substantially worse than the other approaches. This underperformance likely stems from the limited scale of task-specific training data relative to DeBERTa-v3's parameter count, leaving the model insufficient signal to fine-tune its representations effectively for this regression task. Deep learning architectures typically require substantially larger datasets to learn effective representations for domain-specific tasks, and our constrained culinary dataset may not provide enough supervision for a model of this capacity.

To bypass the limitations of sparse training data, we shifted our paradigm to large language models (LLMs) utilizing few-shot inference. This approach yielded substantial improvements. Among the open-source models deployed locally, Gemma-3-27B achieved the strongest results, with scores of 60.21 for Protein and 48.02 for Saturates. Gemini 2.5 Flash, accessed via API, further surpassed all local models, achieving the highest scores for Fat (54.12) and Saturates (59.73).
These results suggest that LLMs can better handle complex culinary reasoning, possibly by leveraging broad pre-trained knowledge for unit normalization and ingredient disambiguation.

Given that the TF-IDF baseline showed reasonable performance despite its simplicity, we hypothesized that augmenting it with LLM reasoning could offer further gains. As shown in the table, employing Gemini 2.5 Flash as a semantic refinement layer on top of the TF-IDF outputs successfully boosted the baseline's performance across all metrics (e.g., improving Protein from 64.62 to 67.79). This supports the viability of a hybrid approach where TF-IDF captures lexical co-occurrence patterns from seen ingredients, while the LLM corrects predictions where semantic understanding (such as unit normalization or ingredient disambiguation) is required.

\begin{table*}[t] 
    \centering
    \renewcommand\arraystretch{1.2} 
    \begin{tabular}{lcccc}
    \toprule
    \multirow{2}{*}{\textbf{Model Architecture}} & \multicolumn{4}{c}{\textbf{Performance Metrics}} \\
    \cmidrule{2-5} 
    & \textbf{Sugar} & \textbf{Protein} & \textbf{Fat} & \textbf{Saturates}  \\
    \midrule
    \multicolumn{5}{c}{\textit{Encoder-based model}} \\
    DeBERTa-v3                  & 8.08  & 28.25  & 9.05  & 8.44  \\
    \midrule
    \multicolumn{5}{c}{\textit{LLM-based Direct Inference}} \\
    Gemma-3-27B        & 42.43 & 59.17 & 38.35 & 46.41\\
    Gemma-3-27B + GPT-OSS-20B        & \textbf{43.03} & \textbf{59.71} & \textbf{39.21} & \textbf{47.13}\\
    \bottomrule
    \end{tabular}
    \caption{Performance comparison of DeBERTa-v3 and the LLMs on the final test set. Metrics follow EU Regulation 1169/2011 tolerance criteria. Bold indicates best performance per metric.}
    \label{tab:test-scores}
\end{table*}

\subsection{Computational Efficiency and Trade-offs}

To ensure a consistent and standardized evaluation of computational efficiency, all local model inferences---spanning the TF-IDF pipeline, DeBERTa-v3, and the open-source LLMs (GPT-OSS-20B, Gemma-3-27B, and Nemotron-3-Nano-30B)---were executed on a dedicated workstation equipped with a single \textbf{NVIDIA RTX PRO 4500 (Blackwell architecture) GPU}. For the hybrid TF-IDF + Gemini 2.5 Flash pipeline, latency measurements encompass the API network routing and server-side generation time.

While predictive accuracy is crucial, practical deployment in dietary monitoring necessitates a strict evaluation of this inference latency. The rightmost column of Table \ref{tab:comprehensive-results} highlights a stark computational trade-off.

The TF-IDF baseline is exceptionally lightweight, boasting a near-instantaneous inference time of 1 ms per recipe.
While DeBERTa-v3 is also highly efficient (3.58 ms), its poor predictive performance disqualifies it for this specific low-data task. 

Conversely, the superior predictive accuracy of LLMs comes at a significant computational cost, even when accelerated by the Blackwell architecture. The few-shot inference times for the open-source LLMs range from 1.4 seconds (Gemma-3-27B) to 23.7 seconds (Nemotron-3-Nano-30B).
Notably, Gemma-3-27B (27B parameters) achieves faster inference than the smaller GPT-OSS-20B (20B parameters) despite its larger model size. This is attributable to differences in quantization: Gemma-3-27B is deployed with Q4\_K\_M quantization, which enables efficient execution on consumer-grade hardware, whereas GPT-OSS-20B uses MXFP4 quantization, which incurs higher overhead on our hardware configuration.
Similarly, both the Gemini 2.5 Flash direct inference and the TF-IDF + Gemini 2.5 Flash pipeline require approximately 1.0 second per request due to API overhead.

Ultimately, for latency-critical applications, the pure TF-IDF model remains the most practical solution. For accuracy-prioritized scenarios, the Gemini 2.5 Flash-based approaches --- both direct inference and the hybrid TF-IDF + Gemini pipeline --- deliver the highest performance across all nutrient categories while maintaining a competitive latency of 1.0 second per request. When only local GPU resources are available without API access, Gemma-3-27B offers the best trade-off among open-source models at a moderate inference cost of 1.4 seconds.

\section{Final Scores on the Test Set}
Based on the results in Table \ref{tab:comprehensive-results}, we submitted our predictions (Team Name: \textbf{andybox111}) for the final test set using DeBERTa-v3, Gemma-3-27B, and the hybrid predictions from Gemma-3-27B and GPT-OSS-20B.
The hybrid prediction is based on 700 samples from GPT-OSS-20B and the rest from Gemma-3-27B.

Although the TF-IDF baseline achieved competitive validation scores, we prioritized neural and LLM-based systems for official submission.
Similarly, the Gemini 2.5 Flash experiments --- both direct inference and the hybrid TF-IDF + Gemini pipeline --- were conducted after the submission deadline as part of our extended analysis for this paper. Furthermore, as a proprietary cloud API, Gemini 2.5 Flash's outputs may vary across API versions, posing challenges for reproducibility.

The official evaluation results are presented in Table \ref{tab:test-scores}.
Consistent with our validation findings, the LLM-based approaches significantly outperform the encoder-based model.
The hybrid approach (Gemma-3-27B + GPT-OSS-20B) achieved the highest scores across all nutrient categories.
However, its performance is only marginally better than the standalone Gemma-3-27B, likely due to the small number of predictions generated by GPT-OSS-20B.

\section{Conclusion}

In this study, we systematically compared three tiers of approaches --- from lexical matching to encoder-based models to generative LLMs --- for automated recipe nutrient estimation. Our experiments revealed that model complexity alone does not guarantee better performance; rather, the availability of pre-trained world knowledge appeared to be a key factor in this low-resource, domain-specific task.

First, we showed that a well-engineered TF-IDF baseline combined with Ridge Regression can achieve reasonable predictive performance with negligible computational overhead, making it a practical choice for resource-constrained or real-time applications. Second, our experiments highlighted a critical limitation of standard deep learning architectures in specialized domains: models like DeBERTa-v3 may struggle under task-specific data scarcity, leading to severe performance degradation. 

Finally, LLM-based approaches delivered the highest accuracy in our experiments. On the official test set, the open-source Gemma-3-27B and its hybrid with GPT-OSS-20B achieved the best submitted scores, suggesting that pre-trained world knowledge may help with complex culinary reasoning without task-specific fine-tuning. In our extended validation analysis, Gemini 2.5 Flash-based approaches further improved performance (e.g., the hybrid TF-IDF + Gemini pipeline improved Fat from 40.58 to 52.12 over the TF-IDF baseline), though these results should be interpreted with caution as they were not evaluated on the official test set and rely on a proprietary API whose outputs may not be fully reproducible.

However, the superior accuracy of LLMs introduces a tangible trade-off with computational efficiency, presenting distinct deployment choices depending on the application's latency tolerance. Future work will explore model distillation and quantization techniques aiming to transfer the complex reasoning capabilities of LLMs into smaller, faster architectures, thereby achieving an optimal balance of accuracy and real-time efficiency for dietary monitoring systems.

\section{Acknowledgements}
We sincerely thank the reviewers for their valuable comments and constructive suggestions, which helped improve the quality of this work. This work was partially supported by the National Science and Technology Council, Taiwan, under Grant No. NSTC 114-2222-E-182-001-MY2.

\section{Bibliographical References}\label{sec:reference}
      
\bibliographystyle{lrec2026-natbib}
\bibliography{lrec2026-example}

\label{lr:ref}
\bibliographystylelanguageresource{lrec2026-natbib}
\bibliographylanguageresource{languageresource}

\end{document}